\documentclass[11pt]{article}

\usepackage[preprint]{acl}
\usepackage{amsmath}
\usepackage{times}
\usepackage{latexsym}
\usepackage{booktabs}

\usepackage{tabularx}
\usepackage[T1]{fontenc}

\usepackage[utf8]{inputenc}

\usepackage{microtype}

\usepackage{inconsolata}

\usepackage{graphicx}
\usepackage{cleveref}

\title{More Yap Less Meaning: Uncovering Self-Improvement Behavior in SLMs}

\author{
\textbf{Marina Igitkhanian}\textsuperscript{1}
\quad
\textbf{Erik Arakelyan}\textsuperscript{2}
\\
\textsuperscript{1}American University of Armenia
\quad
\textsuperscript{2}NVIDIA
}

\begin{document}
\maketitle
\begin{abstract}
Recently, language models have made rapid progress across various domains and applications. However, their capability for self-improvement, i.e., whether they are adept at recognising and correcting flaws in their own reasoning, remains dubious. In this study, we address this question by constructing a sufficiency test to rigorously examine the self-correction capabilities of small language models (SLMs). We propose a minimal three-step self-correction pipeline that collects initial SLM answers, prompts the same model to generate hints for its incorrect responses given the ground truth, and feeds the model the same question with its own feedback to refine the initial answer. We evaluate a variety of instruction-tuned and reasoning SLMs in this experimental setup on arithmetic and logical reasoning benchmarks. Our findings show that SLMs with injected hint sentences yield only a $4.4\%$ gain over initial question-answering accuracy. Even though the correct answer was provided alongside the model’s incorrect reasoning, the evaluated SLMs fail to understand what was missing in their reasoning and show minimal semantic difference between hints that lead to corrections and ones that do not. Furthermore, our experiments show that longer hints are positively correlated with incorrect final answers, suggesting that longer deliberation on problems can hinder the reasoning process, meaning that SLMs do not necessarily scale in performance with a larger compute budget.
\end{abstract}

\section{Introduction}

\begin{figure*}[t!]
    \centering
    \includegraphics[width=\textwidth]{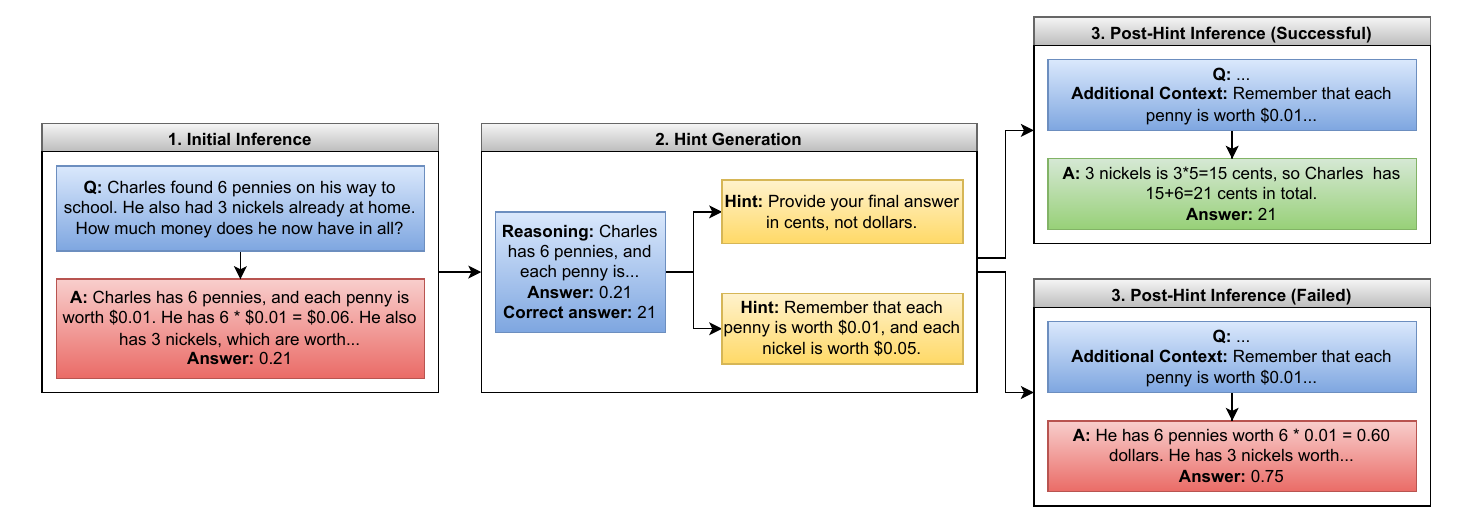}
    \caption{Overview of the three-step self-correction evaluation pipeline.}
    \label{fig:pipeline}
\end{figure*}

Large language models (LLMs) have demonstrated strong performance across a broad range of complex NLP tasks compared to prior state-of-the-art approaches \citep{Kalyan2023GPT3Survey,Wang2018GLUE}. This progress is largely enabled by the attention mechanism in Transformer architectures \citep{Vaswani2017Attention}, large-scale pretraining \citep{Kaplan2020ScalingLaws,Kalyan2023GPT3Survey}, and alignment with human preferences through post-training techniques \citep{Kalyan2023GPT3Survey}. Introduction of Chain-of-Thought prompting has further strengthened LLMs’ ability to solve non-trivial reasoning tasks by generating step-by-step solutions \citep{Wei2022ChainOfThought}. However, LLMs are not perfectly accurate and still produce factual or logical errors \citep{Hadi2023LLMSurvey}. In particular, they are prone to make mistakes in problems, requiring extensive multi-step reasoning, where a single incorrect intermediate step can lead to an incorrect final answer, making failures more likely than in single-step tasks \citep{Miao2023SelfCheck}. To address these limitations, one proposed solution is self-correction, a paradigm in which LLMs improve their initial outputs using feedback generated from their own previous responses \citep{Madaan2023SelfRefine, Miao2023SelfCheck, Gou2023CRITIC, Shinn2023Reflexion, Welleck2023SelfCorrect, Huang2022SelfImprove}. However, other studies question the effectiveness of this approach and suggest that LLMs are not yet capable of reliably identifying their own mistakes and improving their performance \citep{Huang2024CannotSelfCorrect,Jiang2024SelfIncorrect,Valmeekam2023SelfCritique}. Motivated by this uncertainty, we formulate our research question: Are SLMs truly capable of recognising the limitations of their own reasoning and generating feedback to improve their responses? If such awareness exists, a model given the correct answer after an incorrect prediction should be able to identify why its reasoning failed. Consequently, it should then produce feedback or hints that help it arrive at the correct solution. We propose a sufficiency test to evaluate the self-improvement abilities of SLMs through a structured three-step evaluation framework, illustrated in Figure~\ref{fig:pipeline}.

First, the model answers questions from arithmetic and logical reasoning benchmarks, producing chain-of-thought reasoning and final answers. Incorrectly answered questions are then extracted. In the second stage, the model is given its original reasoning, its incorrect answer, and the ground-truth answer, and is asked to generate hint sentences that could help it reach the correct solution on the first attempt. In the final stage, the model reattempts those problems for which hints were generated, using its own feedback as additional context. 
We evaluate the change in accuracy after injecting these hints and perform a qualitative analysis of the hint sentences, examining their length and thematic patterns among both effective and ineffective cases to reveal insights into SLMs' self-improvement capabilities. Towards this end, our contributions are the following: \textbf{(i)} We establish a novel sufficiency test that allows us to rigorously evaluate SLMs for their self-correction capabilities. \textbf{(ii)} We evaluate a variety of SLMs across diverse mathematical and commonsense reasoning domains, showing that SLMs have rather limited capabilities for self-reflection. \textbf{(iii)} We further identify a negative correlation between generated hint lengths and answer correction, showing that SLMs do not necessarily scale in performance with an increase in compute budget.

\section{Related Work}

The term self-improvement can be understood in multiple ways, mainly referring to a process in which an LLM refines its initial response based on the feedback provided after the first generation \citep{Pan2023SelfCorrectionSurvey}. Recent studies distinguish between two types of feedback provided to an LLM: intrinsic and extrinsic. The extrinsic feedback  incorporates inputs from external knowledge sources \citep{Gao2023RARR,Yu2023RetrievalFeedback}, external tools \citep{Gou2023CRITIC,Charalambous2023SelfHealing}, and external evaluation metrics \citep{Jung2022MaieuticPrompting,Welleck2023SelfCorrect}. Relatedly, Reflexion \citep{Shinn2023Reflexion} converts external reward or evaluation signals into verbal reflections that are stored in memory and used to guide subsequent retries. The main limitation of this method is that high-quality data is not always available for most tasks \citep{Madaan2023SelfRefine}. In contrast, intrinsic feedback is generated by the language model itself and does not rely on any external information, and relies solely on the model’s current knowledge. Several studies have explored language model self-improvement using this type of feedback. One is Self-Refine \citep{Madaan2023SelfRefine}, an iterative algorithm that aims to refine the LLM’s initial answer based on feedback given by the same LLM over multiple passes. Similarly, SelfCheck \citep{Miao2023SelfCheck} proposes a zero-shot self-verification approach that regenerates and compares intermediate reasoning steps to detect potential errors without external supervision. Nevertheless, the intrinsic self-improvement method is poorly studied and raises questions in the AI community \citep{Jiang2024SelfIncorrect,Valmeekam2023SelfCritique}. Our setup directly challenges the possibility of purely intrinsic self-correction by framing the task as a sufficiency test for self-improvement. Because the model is given the ground-truth answer, identifying flaws in its initial reasoning should be easier than in a purely intrinsic setting, where no such signal is available. This sufficiency-test framing yields a clean upper-bound argument: failure with oracle access implies failure without it. In other words, if a model cannot detect and address reasoning errors even when the correct answer is provided, it is unlikely to do so reliably when the answer is not available. In addition, prior work on R1-style reasoning models shows that, at some point, the accuracy of the model is negatively correlated with its reasoning length \citep{Marjanovic2026Thoughtology}. Our research shows that the same pattern holds for the self-correction paradigm across instruction-tuned and reasoning SLMs.

\section{Methodology}
\label{methodology}

\paragraph{Sufficiency Test}

Given an SLM $\mathcal{M}$, the answering prompt $\mathcal{P}_{\textit{Ans}} = [T_1, \dots T_{|\mathcal{P}_{\textit{Ans}}|}]$ and the question tokens $\mathcal{Q}_i =[T_1 \dots T_{|\mathcal{Q}_{i}|}]$ from some dataset $\mathcal{Q}_i \in \mathcal{D}$ ,we generate an initial SLM answer in the following way:

$\mathcal{A}^i_{\textit{init}} = \mathcal{M}(\mathcal{P}_\textit{Ans} , \mathcal{Q}_i)
$
Afterwards, we filter out only the samples where the model incorrectly predicted the answer to further test its self-correction, i.e. $\mathcal{D}_{\textit{incorrect}} = \{\mathcal{Q}_i: \mathcal{A}^i_{\textit{init}} \neq y^i_{\textit{gt}}\}$, Where $y^i_{\textit{gt}}$ is the ground truth answer of the question $\mathcal{Q}_i$. Afterwards, we prompt the model $\mathcal{M}$ to generate hint sentences, that does not contain the actual correct answer $\mathcal{H}=[T_1, \dots T_{|\mathcal{H}|}]$, with a specialized prompt $\mathcal{P}_{\textit{hint}} = [T_1, \dots T_{|\mathcal{P}_{\textit{hint}}|}]$ that explicitly contains model's incorrect chain-of-thought, initial incorrect answer $\mathcal{A}^i_{\textit{init}}$ and the ground truth answer $y^i_{\textit{gt}}$ for the designated question $\mathcal{Q}_i$, formalized as $\mathcal{H} = \mathcal{M}(\mathcal{P}_\textit{hint} , \mathcal{Q}_i, \mathcal{A}^i_{\textit{init}})$. If a generated hint contains answer leakage, the hint is rejected and the model is prompted to regenerate it, with a maximum of $3$ attempts. If the model fails to produce an answer-free hint after all attempts, ground-truth answers are masked using the \texttt{<MSK>} token.
 
To finalise our minimal self-correction loop, we add the generated hint sequence $\mathcal{H}$ after the question and generate a final answer in the following way: $\mathcal{A}^i_{\textit{final}} = \mathcal{M}(\mathcal{P}_\textit{Ans}, \mathcal{Q}_i, \mathcal{H})$

\section{Experimental Setup}

\paragraph{Models}
In our experiment, we evaluate a set of open-weight language models from multiple families with sizes ranging from 1.5B to 8B parameters. We select both instruction-tuned (Meta-Llama-3.1-8B-Instruct \citep{grattafiori2024llama3}, Phi-4-mini-instruct \citep{phi42024}, Qwen2.5-Math-1.5B-instruct, Qwen2.5-Math-7B-instruct \citep{yang2024qwen25}, Gemma-2-2B-It \citep{gemmateam2024gemma}), and reasoning-focused (DeepSeek-R1-Distill-Qwen-1.5B, DeepSeek-R1-0528-Qwen3-8B, DeepSeek-R1-Distill-Llama-8B  \citep{deepseekr1}) models. For decoding, we follow the temperature and top-$p$ values recommended for each model family in their official releases. Additionally, we use different token budgets of $256$, $512$, $1024$, $1536$, and $2048$ for the self-correction behaviour.

\paragraph{Benchmarks}
Our test set consists of question-and-answer problems drawn from diverse reasoning benchmarks, including both multiple-choice and free-response formats, and mathematical and commonsense reasoning settings. For numerical reasoning we include GSM8K \citep{Cobbe2021TrainingVerifiers}, ASDiv \citep{Miao2020MathWordProblems}, and AQuA \citep{Wang2017ProgramInduction}. To examine logical and commonsense reasoning, we use AR-LSAT (Analytical Reasoning) \citep{Zhong2021ARLSAT} and the Sports Understanding subset from BIGBench \citep{BIGBench2023BeyondImitation}. More details of the benchmarks can be seen in \cref{sec:benchmarks}.

\paragraph{Evaluation Metrics and Prompts}
\label{subsec:evals}

We evaluate self-improvement via the \emph{accuracy gain} after hint injection.
Given $\mathcal{D}=\{(x_i,y_i)\}_{i=1}^{N}$ with predictions $\hat{y}_i^{(0)}$ (initial) and $\hat{y}_i^{(h)}$ (post-hint), we define

\begin{equation}
\label{eq:acc_gain}
\begin{aligned}
\mathrm{Acc}_{\mathrm{init}} &= \frac{1}{N}\sum_{i=1}^{N}\mathbf{1}\!\left[\hat{y}_i^{(0)}=y_i\right],\\
\mathrm{Acc}_{\mathrm{hint}} &= \frac{1}{N}\sum_{i=1}^{N}\mathbf{1}\!\left[\hat{y}_i^{(h)}=y_i\right],\\
\Delta\mathrm{Acc} &= \mathrm{Acc}_{\mathrm{hint}}-\mathrm{Acc}_{\mathrm{init}}.
\end{aligned}
\end{equation}

Prompting details are in \cref{sec:prompting}.

\section{Results}

\subsection{Self-correction is hard for SLMs}

We evaluate SLMs' self-correction capabilities after receiving self-generated hints across a variety of differing benchmarks and domains. Where available, initial accuracies were consistent with those reported by the model authors, supporting the validity of our generation setup, as shown in Figure~\ref{fig:heatmap_initial_accuracy}.

Across model families, datasets, and token budgets, we observe an average post-hint accuracy gain of $4.4\%$, as can be seen in Figure~\ref{fig:hint_correction_sankey}. A more detailed breakdown is shown in \cref{sec:appendix}.

The results show that SLMs have limited self-correction capacity, suggesting that hint sentences were ineffective and did not capture the information needed to refine their reasoning. Manual evaluation of the hint sequences showed that they were either too general, contained logical inconsistencies, or lacked meaningful content. In particular, specialist models such as Qwen2.5-Math-1.5B-instruct and Qwen2.5-Math-7B-instruct often hallucinated problem continuations and repeated tokens. On average, instruction-tuned and reasoning models yield equal accuracy gains, suggesting that self-refinement capabilities are not associated with the model type, as seen in Figure~\ref{fig:model-category-comparison}.

\subsection{More Yap Less Content}
\begin{figure}[t]
  \includegraphics[width=\columnwidth]{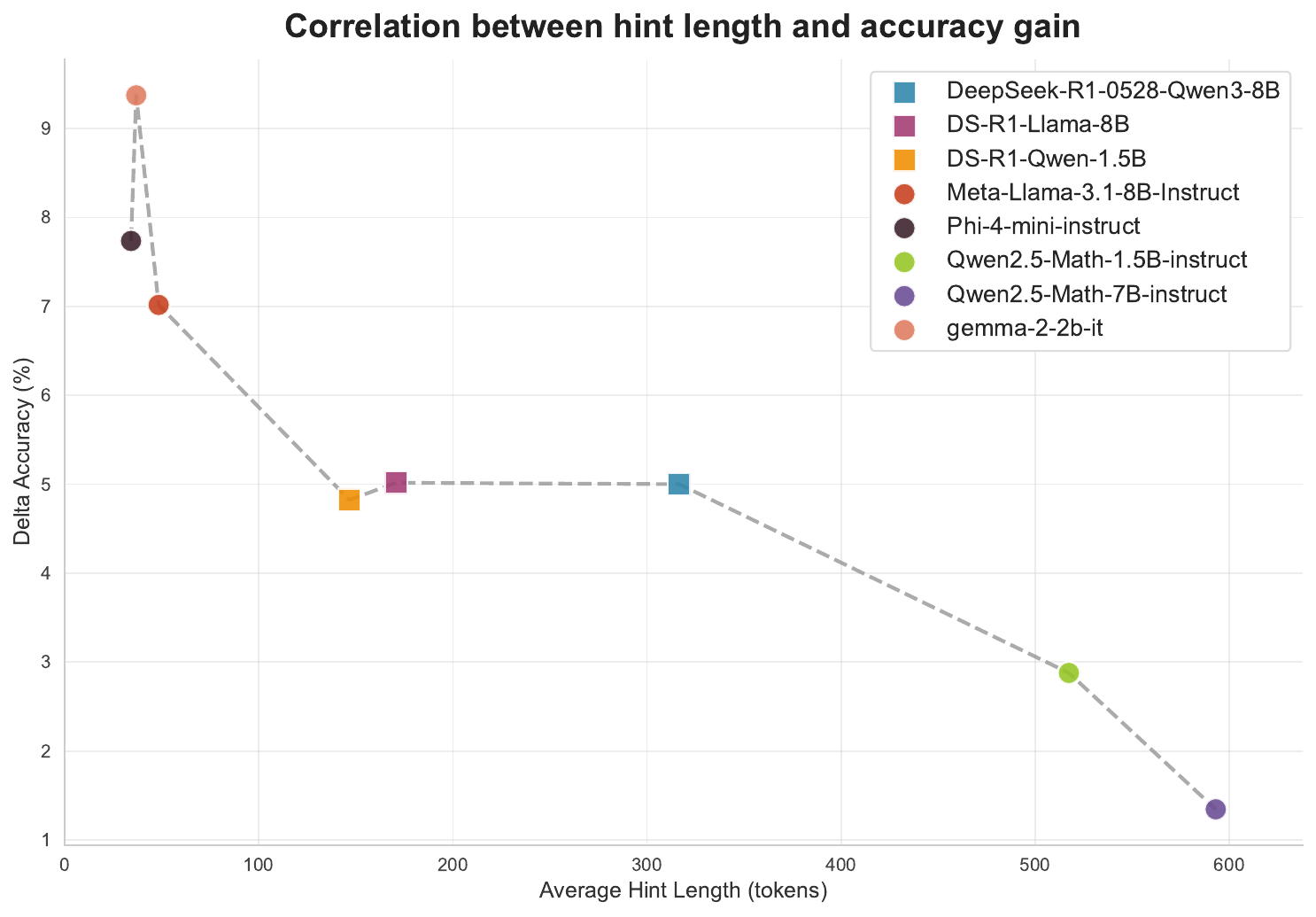}
  \caption{Relationship between average hint length and post-hint accuracy gain across all evaluated models and datasets. Colors correspond to individual models, and marker shapes distinguish model categories, with circles representing instruction-tuned models and squares denoting reasoning models. The horizontal axis shows the average number of generated hint tokens, while the vertical axis reports the change in accuracy after hint injection ($\Delta$ Accuracy).}
  \label{fig:hint-length-correlation}
\end{figure}

We analyze the relationship between hint length and model performance by comparing the number of generated reasoning tokens with post-hint accuracy gain. Our findings, as shown in Figure~\ref{fig:hint-length-correlation}, indicate that hint token length is negatively correlated with post-hint accuracy gain: the longer the hints are on average for a dataset–model pair, the lower the post-hint accuracy gain for that pair. Qualitative analysis of hints shows that in longer hints, SLMs start to hallucinate, often producing repetitive reasoning fragments, instruction leakage ("let me think" or formatting rules), unrelated logical statements, or continuing the problem text beyond the required context. This suggests that if the model is not capable of understanding its reasoning flaw at first, it is unlikely to do so after longer generations. The results support prior observations that extended generation can introduce noise rather than improve solution quality. To address the possibility that this negative correlation is driven by problem difficulty, we manually sample 100 examples with similar question difficulty and observe that both long and short hints occur in comparable cases, with longer hints being mostly noisy and unsuccessful, while short and targeted hints lead to correct final predictions.

\subsection{Useful and useless hints are semantically similar}
To understand what distinguishes successful hints from those that do not capture useful information for the model, we employ the KeyNMF algorithm \citep{kristensenmclachlan2024keynmf} with the embedding model paraphrase-MiniLM-L3-v2
\citep{reimers2019sentencebert, wang2020minilm} to extract representative keywords and latent semantic topics from the generated hints, aggregated across the results from all datasets. This analysis in \cref{tab:semantic} shows that there exists only a minimal semantic difference between successful and ineffective hints. Hence, classifying corrective feedback based solely on lexical heuristics or predefined reasoning and hinting patterns is complicated, highlighting the intricacy of the self-correction task.

\begin{table}[t]
  \centering
  \small
  \begin{tabular}{lr}
    \hline
    \textbf{Correct} & \textbf{\%} \\
    \hline
    model reasoning   & 16.1 \\
    total cost        & 16.1 \\
    basketball player & 12.9 \\
    hockey player     & 12.9 \\
    total distance    & 12.9 \\
    answer boxed      &  9.7 \\
    calculate total   &  9.7 \\
    cookies           &  9.7 \\
    \hline
  \end{tabular}%
  \hspace{0.02\columnwidth}%
  \begin{tabular}{lr}
    \hline
    \textbf{Incorrect} & \textbf{\%} \\
    \hline
    hockey player     & 15.4 \\
    model reasoning   & 15.4 \\
    answer boxed      & 12.8 \\
    basketball        & 12.8 \\
    total cost        & 12.8 \\
    basketball player & 10.3 \\
    calculate total   & 10.3 \\
    help model        & 10.3 \\
    \hline
  \end{tabular}
  \caption{Semantic similarity (top keywords) of hint sentences that resulted in correct or incorrect final predictions.}
  \label{tab:semantic}
\end{table}

\section{Conclusion}
We introduced a three-stage sufficiency test pipeline that incorporates both intrinsic and extrinsic approaches: the model is given the ground truth answer during the hint-generation stage (extrinsic), while the final self-improvement relies on the feedback generated by the model. Our results suggest that current SLMs struggle to generate feedback that improves their predictions, even when provided with ground-truth answers. Failure in this setup allows for stronger conclusions about the limitations of self-improvement. This indicates that SLMs are not yet able to reliably identify the limitations of their own reasoning. Furthermore, our analysis reveals that increased token length in generated hints is frequently correlated with poor outcomes. This observation contributes to prior work that longer outputs are not inherently more informative and may even lead to worse performance.

\section*{Limitations}

Our study has several limitations. Although we evaluate multiple open-weight models from different families and parameter scales, our analysis is restricted to relatively small models (1.5B--8B). Therefore, our conclusions should be interpreted as evidence about self-improvement in SLMs, rather than as a definitive statement about all frontier-scale systems. In addition, although we attempted to control answer leakage by rejecting hints that explicitly reveal the solution and by masking the ground truth when necessary, leakage detection is imperfect. Subtle semantic leakage may still remain, while some harmless hints may be rejected by conservative filtering. This means that the hint-generation stage may still contain artifacts introduced by the prompt design and filtering procedure.

\bibliography{custom}

\appendix

\section{Appendix}
\label{sec:appendix}
\begin{figure}[t]
  \includegraphics[width=\columnwidth]{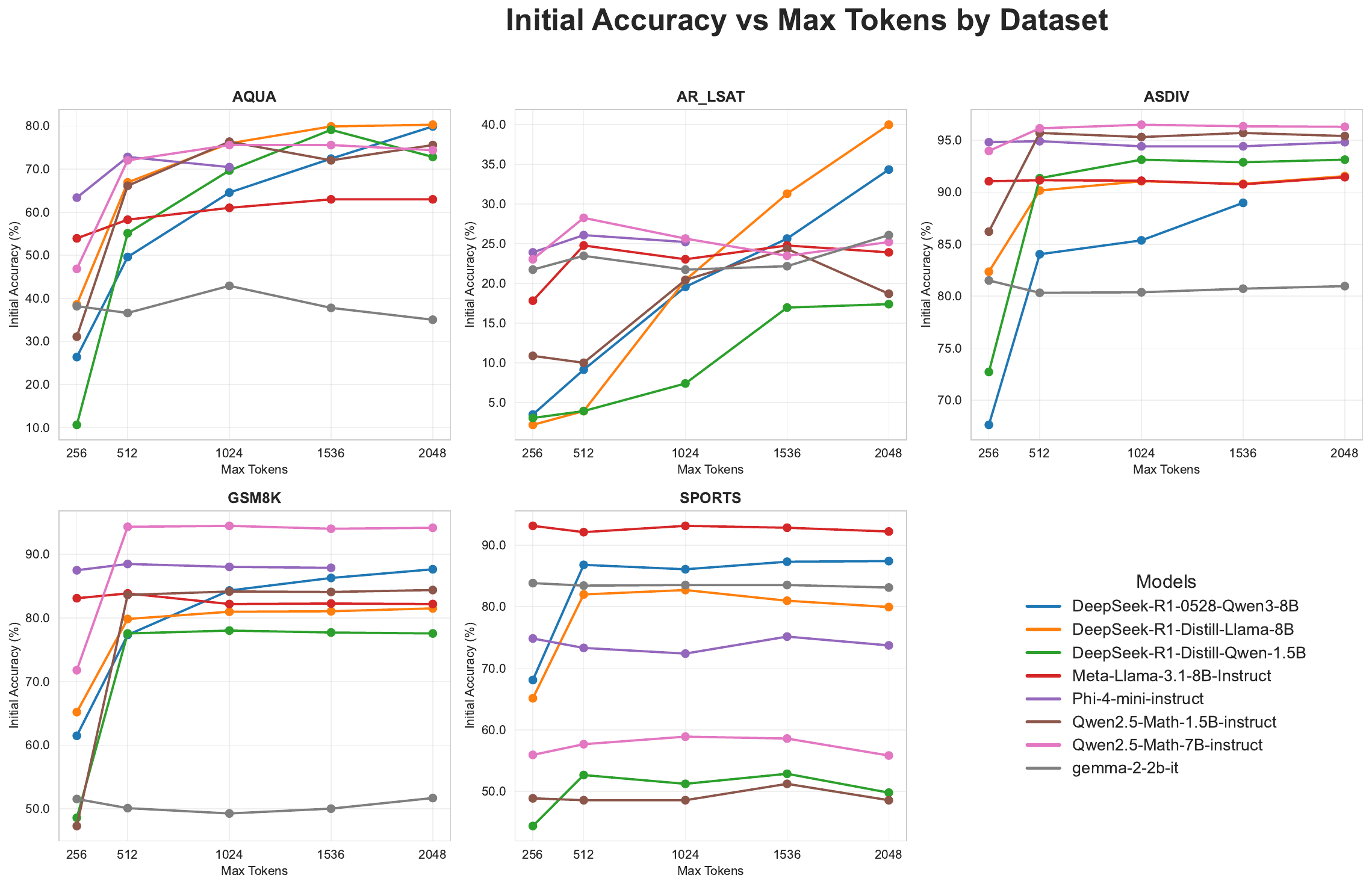}
  \caption{Initial accuracies of all of the models across different tasks and token budgets.}
  \label{fig:tokens_budget_initial_accuracy_by_dataset}
\end{figure}

\begin{figure}[t]
  \includegraphics[width=\columnwidth]{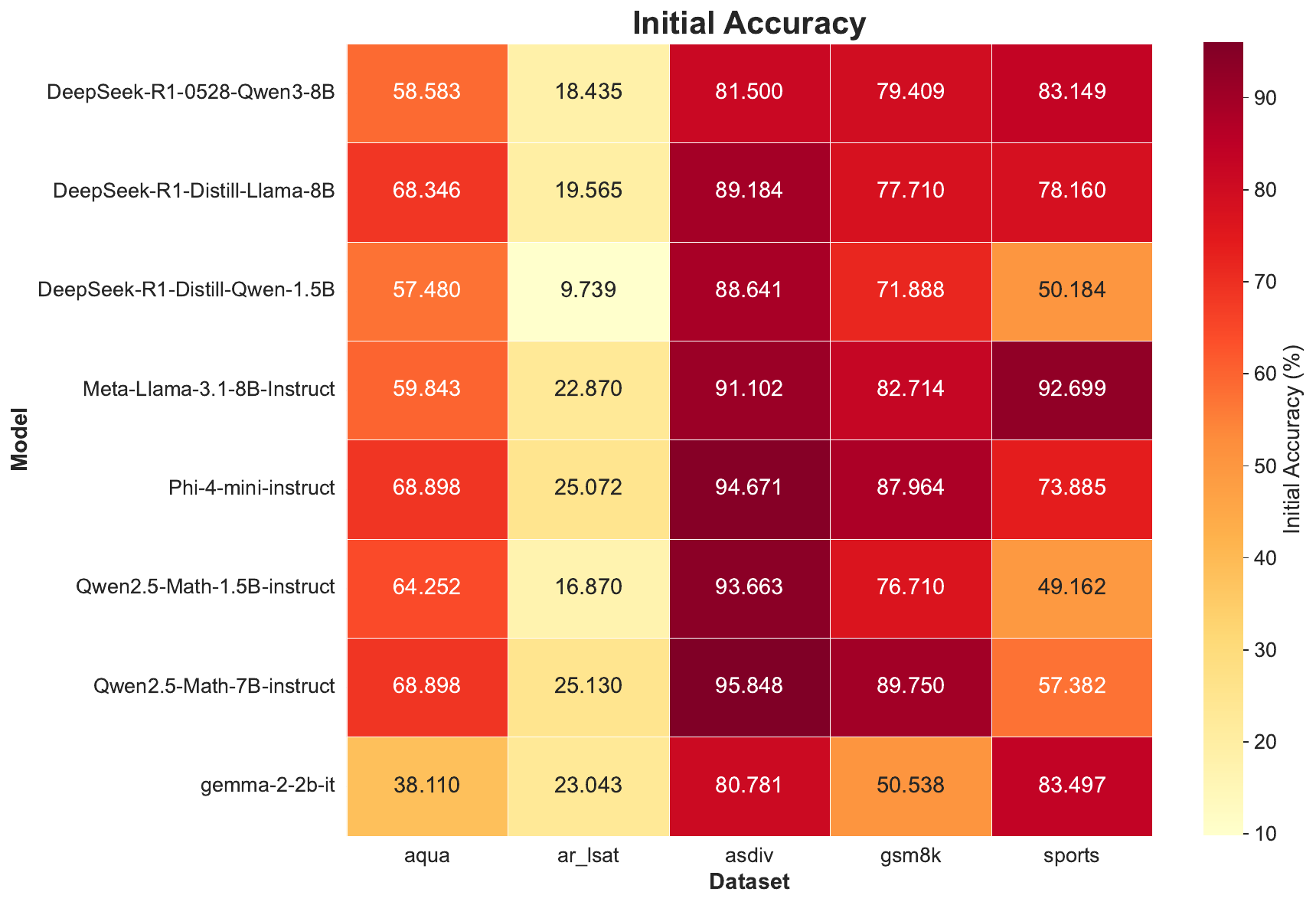}
  \caption{Initial Accuracy (\%) Across Models and Datasets}
  \label{fig:heatmap_initial_accuracy}
\end{figure}

We report the initial accuracies of all of the models across different datasets and token budgets in Figure~\ref{fig:tokens_budget_initial_accuracy_by_dataset} and Figure~\ref{fig:heatmap_initial_accuracy}. Our initial accuracies match the numbers reported by the respective models, thus serving as a sanity check for our study.

\begin{figure}[t]
  \includegraphics[width=\columnwidth]{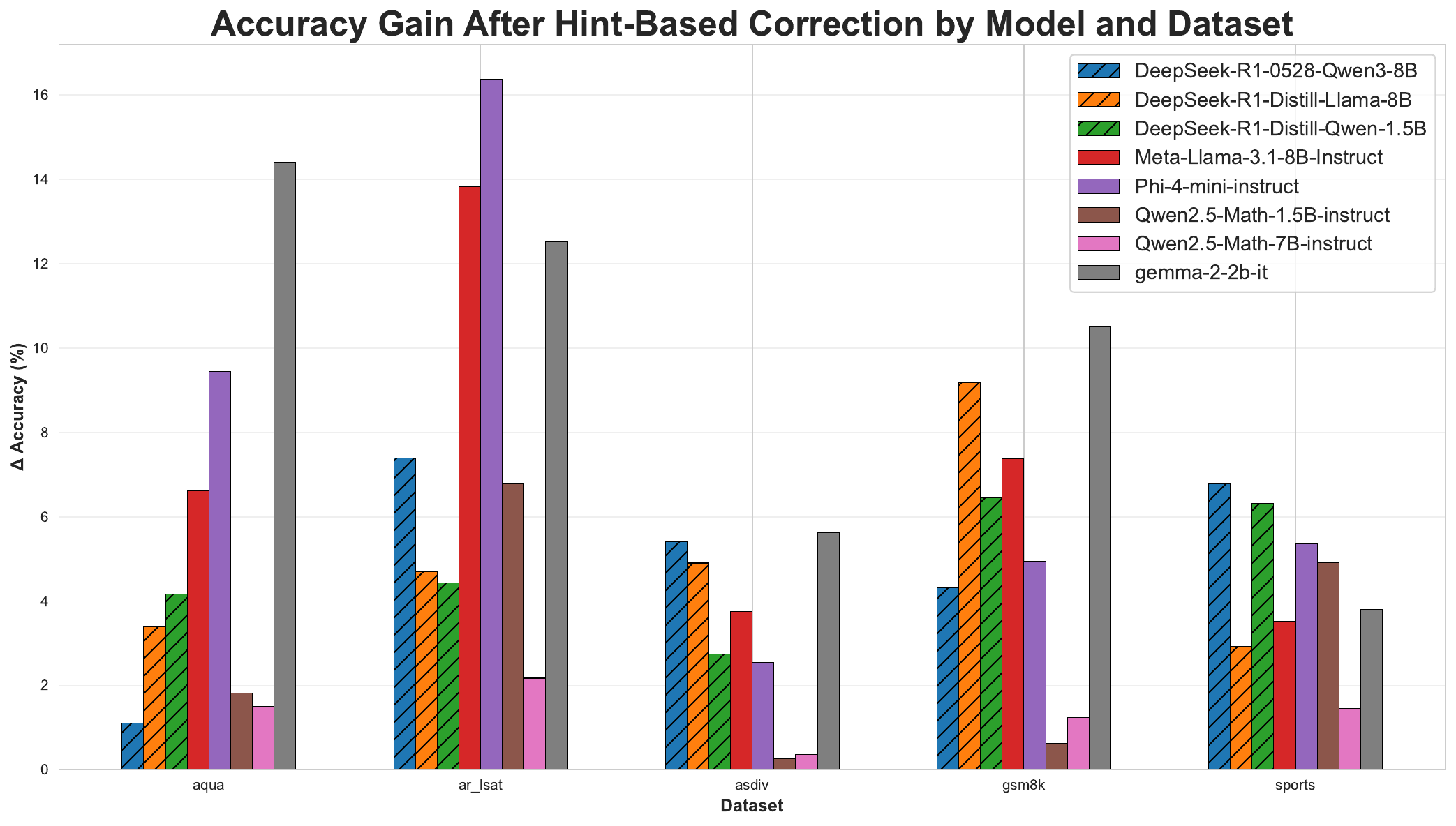}
  \caption{Post-hint accuracy gain ($\Delta$ Accuracy) aggregated across token budgets, shown per model and dataset.}
  \label{fig:aggregated_delta_accuracy_by_dataset}
\end{figure}

Additionally, we report the accuracy gains after hint-based correction across all evaluated models and datasets in Figure~\ref{fig:aggregated_delta_accuracy_by_dataset}.

\begin{figure}[t]
  \includegraphics[width=\columnwidth]{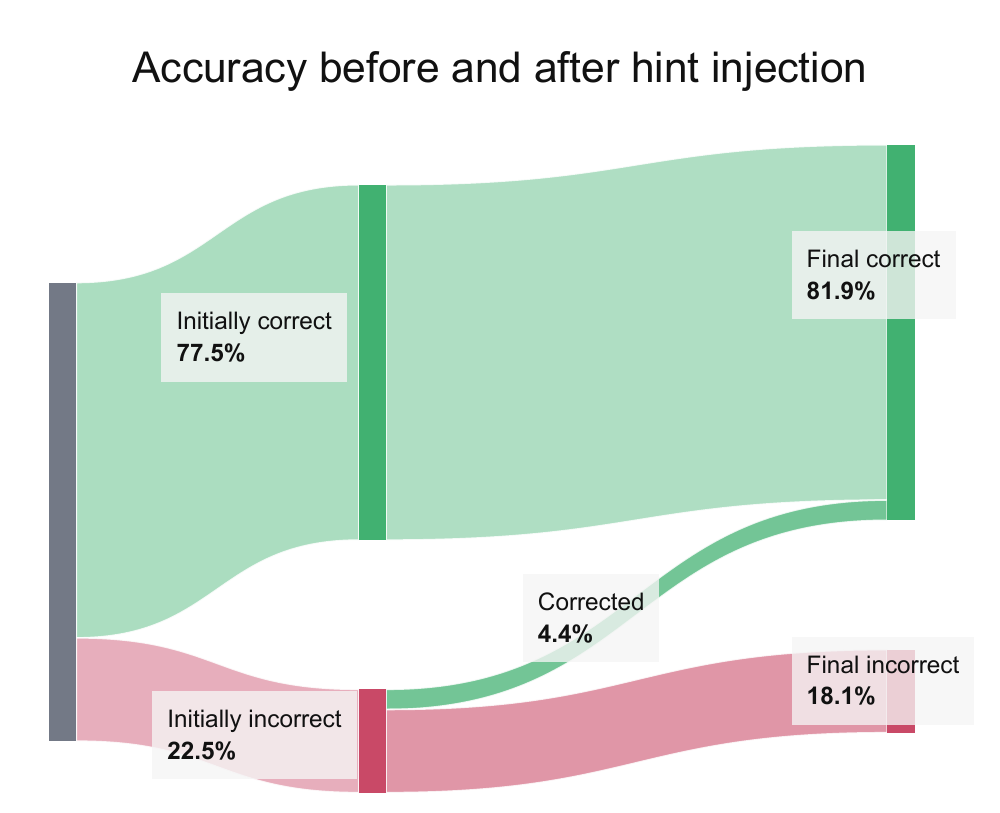}
  \caption{Correction percentages after hint injection averaged across all of the models, tasks and token budgets.}
  \label{fig:hint_correction_sankey}
\end{figure}

\begin{figure}[t]
  \includegraphics[width=\columnwidth]{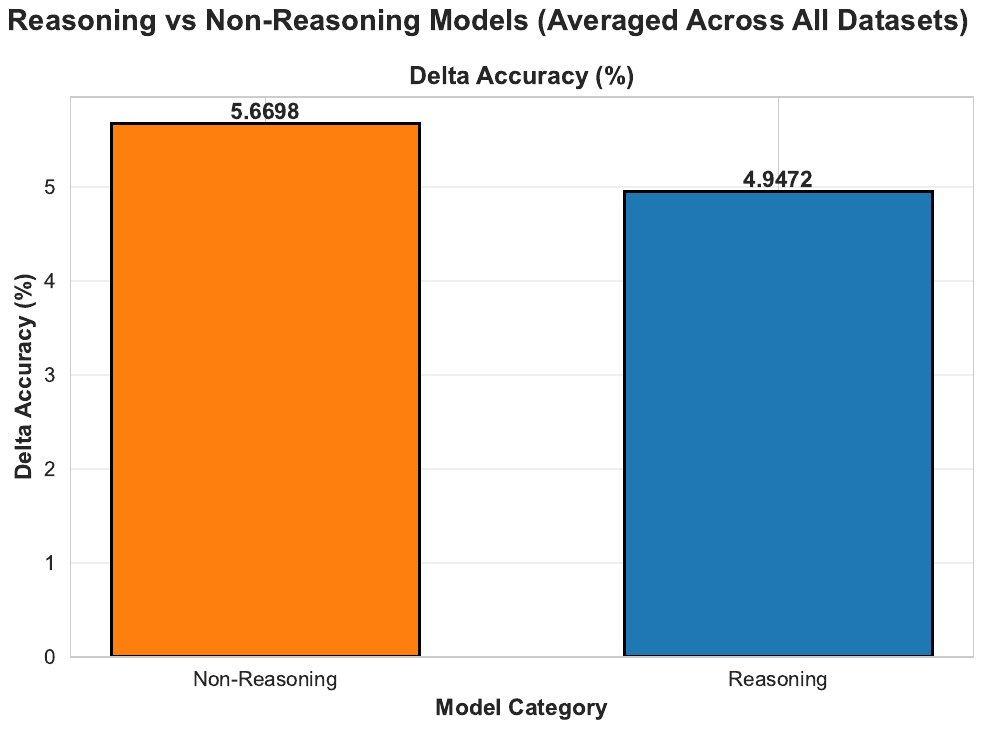}
  \caption{Delta Accuracy of reasoning vs non-reasoning models averaged across all of the datasets.}
  \label{fig:model-category-comparison}
\end{figure}

\label{sec:retry-ablation}

We additionally evaluate a minimal retry baseline to check whether improvements can arise from simply answering the same question again, rather than from the generated hint. In this setting, the generated hint is replaced with a minimal retry instruction stating that the previous answer was incorrect and asking the model to return the final answer again. As shown in \cref{tab:retry-ablation}, this baseline yields small and inconsistent gains across evaluated model--dataset pairs, with an average accuracy gain of only $2.45\%$. Moreover, these gains partly reflect incidental answer changes rather than meaningful self-correction, especially in multiple-choice and binary settings where being told that the previous answer was incorrect substantially narrows the answer space. In binary settings, this effect is especially strong: once the model is told that its previous answer is incorrect, the opposite label should be easily implied. Overall, this suggests that simple retry prompting provides a weak baseline and does not lead to reliable self-correction across tasks.

\begin{table}[t]
  \centering
  \scriptsize
  \setlength{\tabcolsep}{2.2pt}
  \renewcommand{\arraystretch}{1.08}
  \begin{tabular}{lccccc}
    \hline
    \textbf{Model} & \textbf{AQuA} & \textbf{AR-LSAT} & \textbf{ASDiv} & \textbf{GSM8K} & \textbf{Sports} \\
    \hline
    DS-R1-Qwen3-8B    & +0.00 & --     & +0.84 & +0.83 & +0.61 \\
    DS-R1-Llama-8B    & +0.39 & +0.00  & +0.89 & +1.36 & +1.94 \\
    DS-R1-Qwen-1.5B   & --    & +0.00  & +1.09 & +1.06 & +11.04 \\
    Llama-3.1-8B      & --    & --     & +2.52 & +2.65 & +5.93 \\
    Phi-4-mini        & +3.94 & --     & +0.44 & --    & +3.68 \\
    Qwen2.5-Math-1.5B & +0.00 & +0.43  & +0.15 & +0.15 & +3.17 \\
    Qwen2.5-Math-7B   & --    & --     & +0.20 & --    & +1.23 \\
    Gemma-2-2B        & +5.91 & +13.91 & +0.69 & +0.23 & +10.74 \\
    \hline
  \end{tabular}
\caption{Accuracy gain in percentage points for the minimal retry baseline, where the generated hint is replaced with a minimal instruction stating that the previous answer was incorrect and asking the model to return only the final answer. Dashes indicate settings not evaluated.}
  \label{tab:retry-ablation}
\end{table}

\section{Benchmarks}
\label{sec:benchmarks}
GSM8K and ASDiv consist of diverse grade school math word problems, which have a single canonical final answer. AQuA includes multiple-choice math questions covering a broad range of topics and difficulty levels. AR-LSAT (Analytical Reasoning) \citep{Zhong2021ARLSAT} is a dataset of analytical reasoning questions collected from LSAT exams (1991–2016).
It consists of logic problems such as assignment, ordering, rule-based constraints, and grouping. The Sports Understanding subset from BIGBench \citep{BIGBench2023BeyondImitation} assesses the model’s ability to distinguish between plausible and implausible statements related to sports.

\section{Prompting}
\label{sec:prompting}
During the initial and Post-hint inference stages, we use the Chain of Thought prompting technique to encourage explicit reasoning and to obtain intermediate steps required for subsequent hint generation stages. We use dataset-specific prompt templates with unified formatting rules to ensure consistent results parsing which can be found in \cref{sec:appendix}. In the initial and post-hint inference stages, to retrieve model’s chain-of-thought and final answer correctly, those parts of the output were prompted to be included in special tags. To help models follow the required format, we employ a few-shot prompting approach, demonstrating six examples during the initial and post-hint inference stages and four examples during hint generation. This setup ensures stable reasoning behavior while maintaining consistent output structure across datasets. We provide the exact prompts used at all stages of the pipeline. Variables enclosed in \{\} denote placeholders filled at runtime.

\subsection{Initial and Post-Hint Answer Generation Prompts}
\label{subsec:prompts-answer}

We use dataset-specific answering prompts to standardize model outputs and ensure reliable parsing across all models. Each prompt enforces a two-block structure: a reasoning block delimited by \texttt{<think>...</think>} and a final answer block delimited by \texttt{<ans>...</ans>}. Variables enclosed in \{\} denote placeholders filled at runtime. For the post-hint setting, we append the generated hint to the original question while keeping the same output constraints; this isolates the effect of hint injection from formatting differences.

\paragraph{AR-LSAT}
AR-LSAT is evaluated in a multiple-choice format; therefore, we require the final output to be a single option letter (A--E) and forbid any restatement of the question/options to avoid leakage into the answer field (\cref{tab:prompt-template-segmented-noexamples}).

\paragraph{ASDiv and GSM8K}
ASDiv and GSM8K require numeric answers. We enforce a strict numeric-only \texttt{<ans>} field (no units, no words) to prevent spurious formatting variations and to keep answer extraction consistent across models (\cref{tab:mathwp-template-segmented-noexamples}).

\paragraph{AQuA}
AQuA is also multiple-choice. We apply the same constrained letter-only answer format as AR-LSAT, again ensuring that the final line contains only the option letter for robust evaluation (\cref{tab:mcq-template-segmented-noexamples}).

\paragraph{Sports}
For the sports plausibility task, the target label is binary. We map plausibility to \{0,1\} in the \texttt{<ans>} block and explicitly state the decision rule in the prompt to minimize ambiguity (\cref{tab:sports-plausibility-template-segmented-noexamples}).

\subsection{Hint Generation Prompts}
\label{subsec:prompts-hints}

Hints are generated using the model’s original reasoning, its predicted (incorrect) answer, and the ground-truth answer. The hint prompts are designed to provide \emph{helpful guidance without revealing the correct answer}: we disallow explicit mention of the correct option/number and enforce short outputs (1--3 sentences) wrapped in a single \texttt{<hint>...</hint>} block. This design encourages minimal, targeted interventions that can be injected back into the answering prompt while avoiding label leakage.

\paragraph{AR-LSAT}
For AR-LSAT, the hint must not reference option letters or uniquely identify one choice; it should instead supply neutral guidance that could correct the observed reasoning mistake without giving away the answer (\cref{tab:hintgen-template-segmented-noexamples}).

\paragraph{ASDiv and GSM8K}
For numeric word problems, we additionally prohibit directly computing or stating the final numerical result. Hints focus on identifying the correct quantities, operations, and any missing constraints or conversions that led to the initial error (\cref{tab:hintgen-numeric-template-segmented-noexamples}).

\paragraph{AQuA}
For AQuA, we use the multiple-choice hint template that avoids option letters and avoids stating the correct value explicitly, while still pointing the model toward the correct reasoning step (\cref{tab:hintgen-mcq-template-segmented-noexamples}).

\paragraph{Sports}
For sports plausibility, we constrain hints to neutral background knowledge (e.g., the athlete’s sport/position and the relevant meaning of sport-specific terms) and forbid verdict language such as “plausible/implausible” to reduce answer leakage (\cref{tab:hintgen-sports-template-segmented-noexamples}).

\subsection{Initial and Post-Hint Answer Generation Prompts}


\begin{table}[!t]
\centering
\renewcommand{\arraystretch}{1.15}
\footnotesize
\begin{tabularx}{\columnwidth}{p{0.24\columnwidth} X}
\toprule
\textbf{Section} & \textbf{Content (verbatim-style, wrapped)} \\
\midrule

Goal &
{\ttfamily\raggedright
Your goal is to answer the multiple-choice question by showing step-by-step reasoning between <think> and </think> and writing the final answer choice between <ans> </ans>.
\par}
\\

Required output format &
{\ttfamily\raggedright
You MUST output exactly two blocks, in this order, with no extra text before, between, or after them:\par
<think>step-by-step reasoning</think>\par
<ans>X</ans>
\par}
\\

Rules &
{\ttfamily\raggedright
Rules:\par
- X must be exactly one uppercase letter: A, B, C, D, or E.\par
- Do NOT repeat the question, options, or any example text.\par
- Do NOT output anything else after stating the final answer.
\par}
\\

Anti-echo constraint &
{\ttfamily\raggedright
The following are examples of correctly formatted outputs, you must NOT echo those examples or continue providing your own examples.
\par}
\\

Examples &
{\ttfamily\raggedright
(Excluded from this table for brevity; see codebase / full prompt listing.)
\par}
\\

Final instruction + placeholder &
{\ttfamily\raggedright
----\par
Answer the following single question. On the last line return only the correct option letter (A--E). Do NOT output anything else after stating the final answer to this question.\par
\par
Question: \{question\}
\par}
\\

\bottomrule
\end{tabularx}
\caption{Instruction template segmented into components (examples excluded).}
\label{tab:prompt-template-segmented-noexamples}
\end{table}


\begin{table}[t]
\centering
\renewcommand{\arraystretch}{1.15}
\footnotesize
\begin{tabularx}{\columnwidth}{p{0.24\columnwidth} X}
\toprule
\textbf{Section} & \textbf{Content (verbatim-style, wrapped)} \\
\midrule

Goal &
{\ttfamily\raggedright
Your goal is to solve math word problem by showing your step-by-step reasoning between <think> and </think> and writing the final numeric answer between <ans> </ans>.
\par}
\\

Required output format &
{\ttfamily\raggedright
You MUST output exactly two blocks, in this order, with no extra text before, between, or after them:\par
<think>step-by-step reasoning</think>\par
<ans>X</ans>
\par}
\\

Rules &
{\ttfamily\raggedright
Rules:\par
- X must be the final numeric answer with no words and no units. Only number.\par
- Do NOT repeat the question or any example text.\par
- Do NOT output anything else after stating the final answer.
\par}
\\

Anti-echo constraint &
{\ttfamily\raggedright
The following are examples of correctly formatted outputs, you must NOT echo those examples or continue providing your own examples:
\par}
\\

Examples &
{\ttfamily\raggedright
(Excluded from this table for brevity; see codebase / full prompt listing.)
\par}
\\

Final instruction + placeholder &
{\ttfamily\raggedright
----\par
Now answer this single question. Strictly follow the output requirements. Do NOT output anything else after stating the final answer to this question.\par
\par
Question: \{question\}
\par}
\\

\bottomrule
\end{tabularx}
\caption{Math word-problem instruction template segmented into components (examples excluded).}
\label{tab:mathwp-template-segmented-noexamples}
\end{table}


\begin{table}[t]
\centering
\renewcommand{\arraystretch}{1.15}
\footnotesize
\begin{tabularx}{\columnwidth}{p{0.24\columnwidth} X}
\toprule
\textbf{Section} & \textbf{Content (verbatim-style, wrapped)} \\
\midrule

Goal &
{\ttfamily\raggedright
Your goal is to answer the multiple-choice question by showing step-by-step reasoning between <think> and </think> and writing the final answer choice between <ans> </ans>.
\par}
\\

Required output format &
{\ttfamily\raggedright
You MUST output exactly two blocks, in this order, with no extra text before, between, or after them:\par
<think>step-by-step reasoning</think>\par
<ans>X</ans>
\par}
\\

Rules &
{\ttfamily\raggedright
Rules:\par
- X must be exactly one uppercase letter: A, B, C, D, or E.\par
- Do NOT repeat the question, options, or any example text.\par
- Do NOT output anything else after stating the final answer.
\par}
\\

Anti-echo constraint &
{\ttfamily\raggedright
The following are examples of correctly formatted outputs, you must NOT echo those examples or continue providing your own examples.
\par}
\\

Examples &
{\ttfamily\raggedright
(Excluded from this table for brevity; see Appendix / full prompt listing.)
\par}
\\

Final instruction + placeholder &
{\ttfamily\raggedright
----\par
Answer the following single question. On the last line return only the correct option letter (A--E). Do NOT output anything else after stating the final answer to this question.\par
\par
Question: \{question\}
\par}
\\

\bottomrule
\end{tabularx}
\caption{Multiple-choice instruction template segmented into components (examples excluded).}
\label{tab:mcq-template-segmented-noexamples}
\end{table}

\begin{table}[!t]
\centering
\renewcommand{\arraystretch}{1.15}
\footnotesize
\begin{tabularx}{\columnwidth}{p{0.24\columnwidth} X}
\toprule
\textbf{Section} & \textbf{Content (verbatim-style, wrapped)} \\
\midrule

Goal &
{\ttfamily\raggedright
Your goal is to answer given plausibility question by showing step-by-step reasoning between <think> and </think> and writing final answer choice between <ans> </ans>.
\par}
\\

Required output format &
{\ttfamily\raggedright
You MUST output exactly two blocks, in this order, with no extra text before, between, or after them:\par
<think>concise step-by-step reasoning</think>\par
<ans>X</ans>
\par}
\\

Decision rule for X &
{\ttfamily\raggedright
Decision rule for X:\par
- X must be exactly one digit: 1 or 0.\par
- Output 1 if the sentence is reasonably true in a real-world sports setting.\par
- Output 0 if it’s unlikely or inconsistent in a real-world sports setting.
\par}
\\

Rules &
{\ttfamily\raggedright
Rules:\par
- Do NOT repeat the question or any example text.\par
- Do NOT output anything else after stating the final answer.
\par}
\\

Anti-echo constraint &
{\ttfamily\raggedright
The following are examples of correctly formatted outputs, you must NOT echo those examples or continue providing your own examples.
\par}
\\

Examples &
{\ttfamily\raggedright
(Excluded from this table for brevity; see Appendix / full prompt listing.)
\par}
\\

Final instruction + placeholder &
{\ttfamily\raggedright
----\par
Now answer this single question. Strictly follow the output requirements. Do NOT output anything else after stating the final answer to this question.\par
\par
Question: \{question\}
\par}
\\

\bottomrule
\end{tabularx}
\caption{Sports plausibility instruction template segmented into components (examples excluded).}
\label{tab:sports-plausibility-template-segmented-noexamples}
\end{table}

\begin{table}[!t]
\centering
\renewcommand{\arraystretch}{1.15}
\footnotesize
\begin{tabularx}{\columnwidth}{p{0.24\columnwidth} X}
\toprule
\textbf{Section} & \textbf{Content (verbatim-style, wrapped)} \\
\midrule

Task &
{\ttfamily\raggedright
Your job is to look at the model’s reasoning, the model’s chosen answer, and the correct answer, then write a short HINT that would have helped the model arrive at the correct option if that hint had been included in the original question.
\par}
\\

Hint constraints &
{\ttfamily\raggedright
IMPORTANT RULES FOR THE HINT:\par
- Do NOT include the correct answer, the correct option letter, or any wording that points to one unique option.\par
- Do NOT mention any option letters (A--E) or say any option is wrong/right.\par
- The hint must read like neutral background information or guidance, not a verdict.
\par}
\\

Examples &
{\ttfamily\raggedright
(Excluded from this table for brevity; see Appendix / full prompt listing.)
\par}
\\

Generation instruction &
{\ttfamily\raggedright
----\par
Now, based on the reasoning, answer, and the correct answer, write a hint to this question.
\par}
\\

Inputs / placeholders &
{\ttfamily\raggedright
Question: \{question\}\par
Reasoning: \{chain\_of\_thought\}\par
Predicted incorrect answer: \{predicted\_answer\}\par
Correct answer: \{correct\_answer\}
\par}
\\

Output format requirements &
{\ttfamily\raggedright
Return ONLY 1--3 hint sentences, wrapped inside a single <hint>...</hint> block, with no extra commentary and WITHOUT stating correct answer. Do NOT state the correct option letter.
\par}
\\

\bottomrule
\end{tabularx}
\caption{Hint-generation instruction template segmented into components (examples excluded).}
\label{tab:hintgen-template-segmented-noexamples}
\end{table}


\begin{table}[p]
\centering
\renewcommand{\arraystretch}{1.15}
\footnotesize
\begin{tabularx}{\columnwidth}{p{0.20\columnwidth} X}
\toprule
\textbf{Section} & \textbf{Content (verbatim-style, wrapped)} \\
\midrule

Task &
{\ttfamily\raggedright
Your job is to look at the model’s reasoning, the model’s calculated answer, and the correct answer, then write a short HINT that would have helped the model arrive at the correct answer if that hint had been included in the original question.
\par}
\\

Hint format constraints &
{\ttfamily\raggedright
IMPORTANT FORMAT RULES FOR THE HINT:\par
- Do NOT say whether anything is correct or incorrect.\par
- Do NOT write the correct answer in the hint sentences.\par
- Do NOT mention "the correct answer", "your previous answer", or phrases like "you should answer ...".\par
- Do NOT directly compute or state the final numerical result.\par
- The hint must read like neutral background information or guidance, not a verdict.
\par}
\\

What to focus on &
{\ttfamily\raggedright
Focus on:\par
- Clarifying what each quantity in the problem represents and how they relate.\par
- Highlighting the correct operations and order of operations needed (e.g., which numbers to add, subtract, multiply, or divide first).\par
- Pointing out important constraints, unit conversions, or intermediate steps that were overlooked or misapplied.
\par}
\\

Examples &
{\ttfamily\raggedright
(Excluded from this table for brevity; see Appendix / full prompt listing.)
\par}
\\

Generation instruction &
{\ttfamily\raggedright
----\par
Now, based on the question, reasoning, predicted answer, and the correct answer, write a new hint.
\par}
\\

Inputs / placeholders &
{\ttfamily\raggedright
Question: \{question\}\par
Reasoning: \{chain\_of\_thought\}\par
Predicted incorrect answer: \{predicted\_answer\}\par
Correct answer: \{correct\_answer\}
\par}
\\

Output format requirements &
{\ttfamily\raggedright
Return ONLY 1--3 hint sentences, wrapped inside a single <hint>...</hint> block, with no extra commentary and WITHOUT stating correct answer. Do NOT try to solve the problem.
\par}
\\

\bottomrule
\end{tabularx}
\caption{Numeric hint-generation instruction template segmented into components (examples excluded).}
\label{tab:hintgen-numeric-template-segmented-noexamples}
\end{table}

\begin{table}[t]
\centering
\renewcommand{\arraystretch}{1.15}
\footnotesize
\begin{tabularx}{\columnwidth}{p{0.24\columnwidth} X}
\toprule
\textbf{Section} & \textbf{Content (verbatim-style, wrapped)} \\
\midrule

Task &
{\ttfamily\raggedright
Your job is to look at the model’s reasoning, the model’s chosen answer, and the correct answer, then write a short HINT that would have helped the model arrive at the correct option if that hint had been included in the original question.
\par}
\\

Hint constraints &
{\ttfamily\raggedright
IMPORTANT RULES FOR THE HINT:\par
- Do NOT write the correct answer or the correct option letter in the hint sentences.\par
- Do NOT directly compute or state the final numerical result.\par
- Do NOT mention any option letters (A--E) or say any option is wrong/right.\par
- The hint must read like neutral background information or guidance.
\par}
\\

Examples &
{\ttfamily\raggedright
(Excluded from this table for brevity; see Appendix / full prompt listing.)
\par}
\\

Generation instruction &
{\ttfamily\raggedright
----\par
Now, based on the reasoning, answer, and the correct answer, write a hint to this question.
\par}
\\

Inputs / placeholders &
{\ttfamily\raggedright
Question: \{question\}\par
Reasoning: \{chain\_of\_thought\}\par
Predicted incorrect answer: \{predicted\_answer\}\par
Correct answer: \{correct\_answer\}
\par}
\\

Output format requirements &
{\ttfamily\raggedright
Return ONLY 1--3 hint sentences, wrapped inside a single <hint>...</hint> block, with no extra commentary and WITHOUT stating correct answer. Do NOT state the correct option letter.
\par}
\\

\bottomrule
\end{tabularx}
\caption{Multiple-choice hint-generation template segmented into components (examples excluded).}
\label{tab:hintgen-mcq-template-segmented-noexamples}
\end{table}

\begin{table}[t]
\centering
\renewcommand{\arraystretch}{1.15}
\footnotesize
\begin{tabularx}{\columnwidth}{p{0.24\columnwidth} X}
\toprule
\textbf{Section} & \textbf{Content (verbatim-style, wrapped)} \\
\midrule

Task &
{\ttfamily\raggedright
Your job is to look at the model’s reasoning, the model’s predicted answer, and the correct answer, then write a short HINT that would have helped the model arrive at the correct answer if that hint had been included in the original question.
\par}
\\

Hint format constraints &
{\ttfamily\raggedright
IMPORTANT FORMAT RULES FOR THE HINT:\par
- Do NOT say whether the sentence is plausible or not.\par
- Do NOT use the words: "plausible", "implausible", "likely", "unlikely".\par
- Do NOT mention "the correct answer", "your previous answer", or "you should answer ...".\par
- The hint must read like neutral background information or guidance, not a verdict.
\par}
\\

What to focus on &
{\ttfamily\raggedright
Focus on:\par
- Clarifying what sport or position the named player is associated with.\par
- Explaining any sport-specific terms or rules that are relevant.
\par}
\\

Output style requirements &
{\ttfamily\raggedright
Return ONLY 1--3 sentences of hint text, wrapped inside a single <hint>...</hint> block, with no bullet points and no extra commentary.
\par}
\\

Examples &
{\ttfamily\raggedright
(Excluded from this table for brevity; see Appendix / full prompt listing.)
\par}
\\

Generation instruction &
{\ttfamily\raggedright
----\par
Now, based on the question, reasoning, predicted answer, and the correct answer, write hint sentences.
\par}
\\

Inputs / placeholders &
{\ttfamily\raggedright
Question: \{question\}\par
Reasoning: \{chain\_of\_thought\}\par
Predicted incorrect answer: \{predicted\_answer\}\par
Correct answer: \{correct\_answer\}
\par}
\\

Final constraint &
{\ttfamily\raggedright
Return ONLY 1--3 hint sentences, wrapped inside a single <hint>...</hint> block, with no extra commentary and WITHOUT stating correct answer. Do NOT state if the sentence is plausible or not.
\par}
\\

\bottomrule
\end{tabularx}
\caption{Sports plausibility hint-generation template segmented into components (examples excluded).}
\label{tab:hintgen-sports-template-segmented-noexamples}
\end{table}

\end{document}